\documentclass[10pt,twocolumn,letterpaper]{article}

\usepackage{cvpr}              

\usepackage{graphicx}
\usepackage{amsmath}
\usepackage{amssymb}
\usepackage{booktabs}
\usepackage{graphicx}
\usepackage{adjustbox}
\usepackage{multirow}
\usepackage{float}
\usepackage{xcolor}         
\usepackage{colortbl}

\usepackage[pagebackref,breaklinks,colorlinks]{hyperref}

\usepackage[capitalize]{cleveref}
\crefname{section}{Sec.}{Secs.}
\Crefname{section}{Section}{Sections}
\Crefname{table}{Table}{Tables}
\crefname{table}{Tab.}{Tabs.}

\def\cvprPaperID{11212} 
\def\confName{CVPR}
\def\confYear{2023}

\begin{document}

\title{Edge Wasserstein Distance Loss for Oriented Object Detection}

\author{
Yuke Zhu, Yumeng Ruan, Zihua Xiong, Sheng Guo\\
{Mybank Technology\hspace{0.5cm}}\\
}

\maketitle

\begin{abstract}
   Regression loss design is an essential topic for oriented object detection. Due to the periodicity of the angle and the ambiguity of width and height definition, traditional L1-distance loss and its variants have been suffered from the metric discontinuity and the square-like problem. As a solution, the distribution based methods show significant advantages by representing oriented boxes as distributions. 
Differing from exploited the Gaussian distribution to get analytical form of distance measure, we propose a novel oriented regression loss, Wasserstein Distance(EWD) loss, to alleviate the square-like problem.
Specifically, for the oriented box(OBox) representation,  we choose a specially-designed distribution whose probability density function is only nonzero over the edges. 
On this basis, we develop Wasserstein distance as the measure. 
Besides, based on the edge representation of OBox, the EWD loss can be generalized to quadrilateral and polynomial regression scenarios. Experiments on multiple popular datasets and different detectors show the effectiveness of the proposed method.
\end{abstract}

\section{Introduction}
\label{sec:intro}
Oriented object detection has been a fundamental task in various vision area, such as aerial images\cite{xia2018dota}\cite{liu2017hrsc}, scene text\cite{karatzas2015icdar}\cite{yao2012msra_tda}, faces\cite{shi2018face}, and retail scenes\cite{chen2020piou},   it aims to precisely  locate the arbitrarily oriented objects in a single image. Although some methods based on improvements of horizontal object detection have been used in oriented object detection \cite{yang2019scrdet}\cite{qian2021modulated}\cite{yang2021rethinking}\cite{yang2021learning}\cite{yang2018position}\cite{liu2017hrsc} , it is a relatively new direction with many unsolved problems.

\begin{figure*}[!tb]
	\centering
	\begin{subfigure}{0.33\linewidth}
			\centering
			\includegraphics[height=3.3cm]{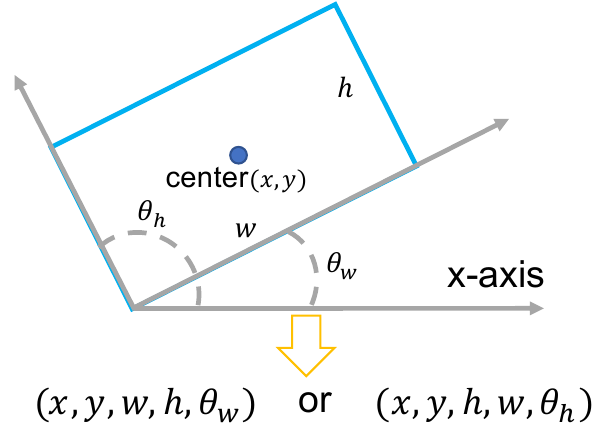}
            \caption{5-p representation}
		\label{fig:v5_rep}
	\end{subfigure}
	\begin{subfigure}{0.3\linewidth}
			\centering
			\includegraphics[height=3cm]{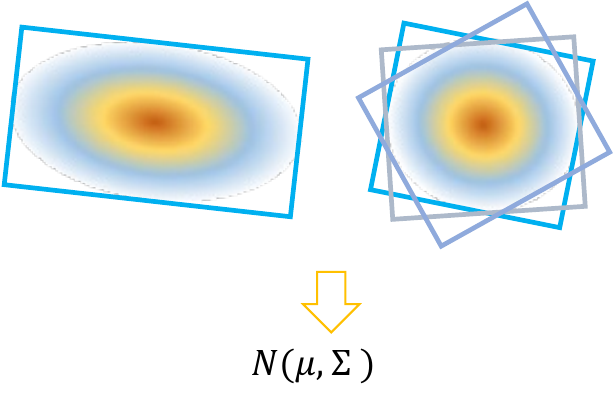}
            \caption{Gaussian representation}
		\label{fig:gaussian_rep}
	\end{subfigure}
	\begin{subfigure}{0.3\linewidth}
			\centering
			\includegraphics[height=3cm]{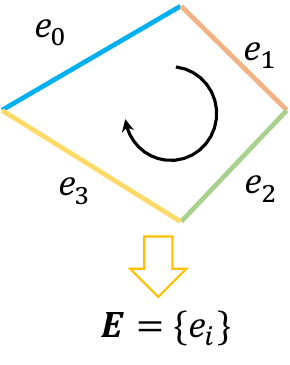}
            \caption{Edge representation}
		\label{fig:edge_rep}
	\end{subfigure}
	\centering
	\caption{Illustration of oriented bounding box representations. a) Using five parameters $(x,y,w,h,\theta)$ to represent a rotated bounding box. A oriented box can be mapped to two variants in the 5-p vector space. b) Gaussian representation proposed by GWD\cite{yang2021rethinking}. It converts the oriented box to Gaussian distribution. However, multiple squares are likely to be mapped to one Gaussian. c) Our proposed edge representation. We use a clockwise(or counter-clockwise) edge sequence to represent the oriented box. It is not affected by the square-like problem.}
	\label{fig:box_rep}
	\vspace{-12pt}
\end{figure*}

Designing a proper regression loss is essential for training a well-performed oriented detector. However, due to the periodicity of the angle and the ambiguity of width and height definition, the regression process has been suffering from metric discontinuity and square-like problems. 
Recently, some works\cite{yang2019scrdet}\cite{qian2021modulated} try to modify the L1 distance loss in order to solve these limitations. They either bring in extra regression term or depends on new representation definition, and the resultant performance is far from satisfactory.
Among all the literature on oriented regression loss design, GWD\cite{yang2021rethinking} and KLD\cite{yang2021learning} are the first to propose representing the OBox as distributions. They convert the OBox to Gaussian distributions and use Wasserstein distance or Kullback-Leibler Divergence as the regression loss. The distributional representation elegantly solves the periodic problem of angle and ambiguity problem of width and height. 
However, 
the Gaussian distribution degenerates for squares and thus fails to capture the angular information as showed in Figure-\ref{fig:gaussian_rep}.Besides, those near square objects will also have trouble in regressing the angle precisely(shown in Figure-\ref{fig:vis_compare}) as their optimization process is near stopping. Therefore, we believe that there should be a much more reasonable choice other than the Gaussian representation to fully exploit the advantage of the distributional representation method. In this paper, we follow the idea of distributional representation and propose a novel oriented regression loss called Edge Wasserstein Distance(EWD) loss to solve all the above problems. 

In order to better reserve the geometrical information of OBox, we introduce  a specially-designed distribution whose probability density function is only nonzero over the edges as its representation. In this way, we develop the Wasserstein distance  as the  loss to regress the distribution between the ground-truth and prediction. As it is difficult to compute this distribution loss directly, we introduce several assumptions to simplify the calculation and present a simple approximation of  Edges Wasserstein distance(EWD) as the regression loss. 
The proposed EWD has several advantages. First, with the distributional representation, it solves all the limitations of previous work, immune from both the angular periodicity and ambiguity of width and height. Also the square-like problem will not exist. Second, by treating the OBox as a whole, the optimization of boxes' parameters are dynamically adjusted according to the chosen target. This leads to high performance especially in high-precision cases. Third, the proposed EWD, along with the edge representation, is more general compared to other distribution based methods. It is applicable to quadrilateral and polynomial regression scenarios. In summary, the contributions of this paper are three folds:

1) We introduce edge representation and propose a novel oriented regression loss, Edge Wasserstein Distance loss, which is simple and general to rectangular, quadrilateral and polynomial regressions.

2) From general to specific, we theoretically develop the formulation of EWD loss on oriented bounding box occasions. The resultant formulation is simple and turns out to be a generalized form of horizontal L2-distance in oriented cases. 

3)  Experimentally, we show that for three datasets and two popular detectors, the EWD loss can achieve challenging results compared with its peer methods.

\section{Background}
\subsection{Related Work}
Oriented object detection is an extension to classical horizontal detection. It is widely used in areas like aerial object detection\cite{he2016resnet}, scene text detection\cite{zhou2017east}\cite{long2018textsnake}\cite{liao2020dbnet} where targets need to be more precisely located. To find the orientation, there are mainly three approaches: the classification based \cite{yang2021dense}\cite{yang2020arbitrary}, segmentation based\cite{wang2019mask}\cite{liao2020dbnet} and regression based methods\cite{zhou2017east}\cite{liu2017rotatedrpn}\cite{jiang2017r2cnn}\cite{qian2021rsdet++}. The latter approach is preferred by most researchers for its simplicity and advantage in getting high-precise localization. However, the oriented regression has been suffered from discontinuity problem and square-like problem when using the traditional L1 loss. These two problems are mainly caused by the periodicity of angle and variations in bounding box definition. 
As a remedy for L1 loss, SCRDet\cite{yang2019scrdet} proposes IoU-Smooth L1 loss to smooth the boundary of discontinuity by introducing the skew-IoU factor. RSDet\cite{qian2021modulated} proposes modulated loss to overcome the discontinuity. Although they partly solve these problems, the performances are far from satisfactory. Other methods try to introduce the IoU loss \cite{yu2016unitbox}\cite{zheng2020diou}\cite{rezatofighi2019giou} in horizontal detection into oriented detection. They usually approximate the IoU loss as its implementation in oriented occasions is too complicated. PolarMask\cite{xie2021polarmask++} proposes an approximation of IoU operation. However, it is discrete and cannot be directly used for back propagation. Similar idea is taken by PIoU\cite{chen2020piou}. The most similar work to our method is GWD\cite{yang2021rethinking} and KLD\cite{yang2021learning}, which both represent the oriented bounding box as Gaussian distribution. However, the square-like problem still exists. Moreover, they are limited to regressing the rectangular boxes and can not be generalized to quadrilateral cases.

\subsection{Oriented Bounding Box Representation}
\label{sec:box_rep}
For rotated bounding box representation, the most common method is the 5-p representation denoted in Figure-\ref{fig:v5_rep}. 
The problem for 5-p representation lies in two aspects. First, the width and height is exchangeable. So when L1 regression loss is used, the regression target is unclear.
Second, the angle is periodic, the regression target for angle is also ambiguous.
To avoid the limitations of 5-p representation, GWD and KLD propose to represent the oriented box as Gaussian distribution $(\mu,\Sigma)$, where $\mu = (x,y)$, $\Sigma^{1/2}=RSR^{\top}$ and $R$ is the rotation matrix defined by $\theta$. The Gaussian representation solves the problem of ambiguity between width and height. On this basis, the boxes' distance can be calculated via popular distributional distance measure, KL-Divergence or Wasserstein distance. 
The distributional representation leads to better performance by treating the oriented box as a whole.
Specifically, the optimization process of each parameter is dependent on each other and adaptively adjusted. 
In contrast, the 5-p representation together with L1 loss treat each parameter independently, thus getting sub-optimal performance. 

Following the distributional representation, we find that Gaussian representation may not be the perfect choice. As illustrated in Figure-\ref{fig:gaussian_rep}, the Gaussian representation degenerates for the squared box as the rotation of squares will not change their Gaussian representation. In other words, Gaussian representation is not a one-to-one mapping. It loses the angular information for squares. 

To better represent the OBox, we design a distribution to capture the full geometrical information, which we call edge representation(shown in Figure-\ref{fig:edge_rep}). 
Specifically, we use a directional edge sequence $\bf E$ to represent a oriented box. 
The direction of the sequence is either clockwise or counter-clockwise. Without loss of generality, we take the clockwise direction.
For the probability distribution, we design a 2-dimensional distribution which is only defined on the edges and zero-valued elsewhere. The probability density function of OBox is written as 
\begin{equation}
    p(x,y) = 
    \begin{cases}
    \Tilde{p}(x,y) & (x,y)\text{ on }\bf{E}\\
    0 & \text{Otherwise}
    \end{cases}
\end{equation}
where $\tilde{p}$ satisfies $\int_{\bf{E}}\tilde{p}(x,y) \,ds = 1$. 
The edge representation is general and can be applied to quadrilateral and polynomial cases. Next, we will develop the distance measure based on the edge representation. 

\section{Proposed Method}
In this section, we introduce the proposed Edge Wasserstein Distance loss(EWD). It is based on the Wasserstein distance of bounding boxes' edge. In the first part, we formulate the distance measure of bounding boxes. In the second part, we design the EWD loss on this basis. In the final part, we analyze and show some good properties of EWD.
\subsection{Edge Wasserstein Distance}

Based on the previous edge representation, we formulate the distance between two OBoxes' distributions, $\bf{p}$, $\bf{q}$ by the Wasserstein metric
\begin{equation}\label{wass}
    \bf{W(p,q)} = \inf \mathbb{E}(||\bf{X} - \bf{Y}||_2^2)
\end{equation}
where $\bf{X}$ and $\bf{Y}$ are points sampled from distribution $\bf{p}$ and $\bf{q}$ respectively.
However, as the definition of Wasserstein distance involves an optimization process, the calculation of Eq-\ref{wass} is complex.
As a simplification, we make an assumption on the joint distribution of two sampled points $\bf{X}$ and $\bf{Y}$. Suppose ${\bf{P}_i}$ and ${\bf{Q}_i}$ are corner points of two OBoxes where $i  \in \{0,1,2,3\}$ and $\bf{X}_i$ and $\bf{Y}_j$ are sampled from segment $\bf{P}_i\bf{P}_{i+1}$ and $\bf{Q}_j\bf{Q}_{j+1}$ respectively. We restrict $p(\bf{X}, \bf{Y})$ by 
\begin{equation}\label{eq:assum1}
    p(\bf{X}_i, \bf{Y}_j) = 
    \begin{cases}
    \Tilde{p}(\bf{X}_i, \bf{Y}_j) & j = \bf{\pi}(i) \\
    0 & \text{otherwise}
    \end{cases}
\end{equation}
where $\bf{\pi}$ denotes a bipartite matching function which maps the edge index $i$ from one OBox to $\mathbf{\pi}(i)$ of the other. 
The meaning of this assumption is that the distance calculation should only be conducted between the geometrical edge counterparts. 
Thus, the \ref{wass} can be simplified as 
\begin{equation}\label{wass2}
\begin{split}
    \bf{W(p,q)} &= \inf_{p(\bf{X_i}, \bf{Y}_j)} \mathbb{E}\sum_i\sum_j(||\bf{X}_i - \bf{Y}_j||_2^2) \\
    &= \inf_{\bf{\pi}}\sum_i(\inf_{p(\bf{X_i}, \bf{Y}_{\bf{\pi}(i)})}\mathbb{E}||\bf{X}_i - \bf{Y}_{\bf{\pi}(i)}||_2^2)
\end{split}
\end{equation}
The meaning of Eq-\ref{wass2} is that distance calculation of the OBoxes is divided into the distance measure of each segment pair $\mathbb{E}||\bf{X}_i - \bf{Y}_{\bf{\pi}(i)}||_2^2$. 
In Eq-\ref{wass2}, the outer infimum aims to find a bipartite matching  $\bf{\pi}$ that minimize the distance.
As the edge sequence is defined to be directional, the total number of match is equal to the edge number. For rotated box, the number is 4.
In practice, we traverse all the 4 matches to get the minimum value as the distance.
Further simplification of Eq-\ref{wass2} depends on either the distribution $\Tilde{p}(x, y)$ of the OBox edge or the joint distribution $\tilde{p}(\bf{X}_i, \bf{Y}_j)$. Here, 
we introduce two approximations to simplify the Eq-\ref{wass2} and formulate the Edge Gaussian Wasserstein Distance(EGWD) and Edge Dense Wasserstein Distance(EDWD).

\noindent\textbf{Edge Gaussian Wasserstein Distance (EGWD)} 
We suppose $\Tilde{p}(x, y)$ to be a 2-D Gaussian distribution denoted by $(\mu, \Sigma)$, where $\mu$ represents the edge's center point and $\mathbf{\Sigma}$ is the covariance. 
The covariance is calculated as $\mathbf{\Sigma}^{1 / 2} =\mathbf{R S R}^{\top}$ where $\mathbf{R}$ is the rotation matrix and $\mathbf{S} = diag(\frac{w}{2}, 0)$ with $w$ being the edge length.
Following the above derivation, it turns out that the Wasserstein distance between two consistent edge pair$(\mu_1, \Sigma_1)$ and $(\mu_2, \Sigma_2)$ is written as\cite{wasserstein2010djalil}:
\begin{equation}\label{wass_gaussian}
    \bf{W_{12}} = ||\mu_1 - \mu_2||_2^2 + \textbf{Tr}\left(\Sigma_1 + \Sigma_2 - 2(\Sigma_1^{1/2}\Sigma_2\Sigma_1^{1/2})^{1/2}\right)
\end{equation}
Thus, the Eq-\ref{wass2} can be written as the summation of each paired distance.

\noindent\textbf{Edge Dense Wasserstein Distance (EDWD)} 
In EDWD, the joint distribution defined in Eq-\ref{eq:assum1} further restricted for simplification.
We further restrict the joint distribution for $\tilde{p}(\bf{X}_i, \bf{Y}_j)$ by Eq-\ref{eq:assum2}. 
\begin{equation}\label{eq:assum2}
    p(\bf{X}_i, \bf{Y}_j) = 
    \begin{cases}
    \Tilde{p}(\bf{X}_i, \bf{Y}_j)) & j = \bf{\pi}(i)\text{ and } \delta(X_i) = \delta(Y_i)\\
    0 & \text{otherwise}
    \end{cases}
\end{equation}
where $\bf{\delta}(\bf{X}_i) = ||\bf{X}_i\bf{P}_i|| / ||\bf{P}_i\bf{P}_{i+1}||$ representing the normalized position of $\bf{X}_i$ on the segment. It means that each point on the edge is matched exactly to its counterparts on the paired edge. As the sampling process of $\bf{X}_i$ and $\bf{Y}_i$ becomes fully coupled,  the joint distribution $\Tilde{p}(\bf{X}_i, \bf{Y}_j)$ degenerates to the edge's distribution, which becomes a one dimensional distribution. Then Eq-\ref{wass2} can be further simplified by a line integration. We directly show the final formulation for Eq-\ref{wass} and leave the derivation to the appendix.
\begin{equation}
    \label{eq:wass_final}
    \begin{split}
    \bf{W(p,q)} = \inf_{\bf{\pi}}\sum_i \left(||\Delta c_i||^2 + \frac{\sigma^2}{4} ||\Delta w_i||^2\right) \\
    \end{split}
\end{equation}
where $\bf c_i$ denotes center of the segment, $\bf{\Delta c_i}$ denotes the vector between paired segments' center points, $\bf {w_i}$ denotes the edge vector and $\bf{\Delta w_i}$ denotes the vector difference. $\sigma ^ 2$ denotes the variance of the designed paired edges' distribution $\Tilde{p}$. The geometrical meaning of this approximation is to force every point sampled from the edge to be densely aligned to its counterparts. Hence, the regression can be viewed as a dense alignment process. 

\subsection{Edge Wasserstein Distance Regression Loss for Oriented Bounding Box}
The developed EGWD and EDWD in Eq-\ref{eq:OBox_egwd} and Eq-\ref{eq:OBox_edwd} are for general polygons.
This property is guaranteed by Eq-\ref{eq:assum1} by which we can approximates the Wasserstein distance by 
decoupling the calculation to the distance measure of each edge pair.

For OBoxes, the edges are not independent. Thus, we can further bring in the  geometrical constraint and develop EWD for OBox.  Without loss of generality, we take the $\theta$-based representation $(x,y, w, h, \theta)$ for OBox.

\noindent \textbf{EGWD for OBox}. 
There are two constraints on the condition of OBox. First, the center point of each edge can be expressed by the 5-p vector $(x,y, w, h, \theta)$. Second, the parallel edges share identical covariance.
By clockwise order, the center points can be written as
\begin{equation}
\begin{split}
    \mathbf{\mu}_1 &=(x - \frac{h}{2}\sin \theta, y - \frac{h}{2}\cos \theta)^{\top}, \\
    \mathbf{\mu}_2 &=(x + \frac{w}{2}\cos \theta, y - \frac{w}{2}\sin \theta)^{\top}, \\
    \mathbf{\mu}_3 &=(x + \frac{h}{2}\sin \theta, y + \frac{h}{2}\cos \theta)^{\top}, \\
    \mathbf{\mu}_4 &=(x - \frac{h}{2}\sin \theta, y + \frac{w}{2}\cos \theta)^{\top}
\end{split}
\end{equation}
And the two covariance matrices that correspond to $w$ and $h$ are written as:
\begin{equation}
\begin{split}
\mathbf{\Sigma}^{1 / 2}_w 
&=\left(\begin{array}{cc}
\frac{w}{2} \cos ^{2} \theta & \frac{w}{2} \cos \theta \sin \theta \\
\frac{w}{2} \cos \theta \sin \theta & \frac{w}{2} \sin ^{2} \theta
\end{array}\right), \\
\mathbf{\Sigma}^{1 / 2}_h 
&=\left(\begin{array}{cc}
\frac{h}{2} \sin ^{2} \theta & \frac{h}{2} \cos \theta \sin \theta \\
\frac{h}{2} \cos \theta \sin \theta & \frac{h}{2} \cos ^{2} \theta
\end{array}\right)
\end{split}
\end{equation}
Then for a given match $\bf{\pi}$, the summation of the former term of Eq-\ref{wass_gaussian} can be simplified into
\begin{equation}\label{eq:simp}
    \sum_i ||\mu_1 - \mu_2||_2^2 =\bf 4||\Delta o||^2 + \frac{1}{2}||\Delta w||^2 + \frac{1}{2}||\Delta h||^2
\end{equation}
where $\bf{\Delta o} = (\Delta x)^2 + (\Delta y) ^ 2$ represents the center point distance and $\bf{\Delta w} = ||(w_1 \cos \theta_1, w_1 \sin \theta_1)^{\top} - (w_2 \cos \theta_2, w_2 \sin \theta_2)^{\top}||^2$ represents the width vector difference. $\bf{\Delta h}$ is similar to $\bf{\Delta w}$. 
Similarly, we develop the simplified expression for the summation of latter term of Eq-\ref{wass_gaussian} as
\begin{equation}\label{eq:trace}
    \textbf{Tr}(...) = \frac{1}{2}||\bf{\Delta w}||^2 + \frac{1}{2}||\bf{\Delta h}||^2
\end{equation}
Thus, we combine Eq-\ref{eq:simp} and Eq-\ref{eq:trace} together and get the formulation for EGWD distance measure as
\begin{equation}\label{eq:OBox_egwd}
    \bf{W_{egwd}(p,q)} =\bf 4||\Delta o||^2 + ||\Delta w||^2 + ||\Delta h||^2
\end{equation}
The full derivation is left to the \textbf{appendix}. 
At first glance, we find that the formulation of EGWD is the generalized form of L2-distance loss.

\noindent \textbf{EDWD for OBox}. The simplification process of EDWD is similar to EGWD as they both contain the summation of edge center difference. On the basis of Eq-\ref{eq:simp}, the Eq-\ref{eq:wass_final} can be written as
\begin{equation}
    \label{eq:OBox_edwd}
    \begin{split}
    \bf{W_{edwd}(p,q)} &=  \bf 4||\Delta o||^2 + \left(\frac{1}{2} + \frac{\sigma_w^2}{2}\right)||\Delta w||^2 + \\ & \left(\frac{1}{2} + \frac{\sigma_h^2}{2}\right)||\Delta h||^2
    \end{split}
\end{equation}
where the $\sigma_w^2$ and $\sigma_h^2$ represent the variances of the single-variate distribution that correspond to the width and the height. Their values describe the geometrical information of the corresponding edges. In this work, we design the variance to be $w$ for width and $h$ for the height. In this way, the variance ratio keeps the OBoxes' aspect ratio. This special design leads to better performance compared to directly setting them to a constant value. 

\subsection{Overall Loss Function Design}

\noindent\textbf{Relationship between EGWD and EDWD}.
EGWD and EDWD have similar formulations.
Comparing Eq-\ref{eq:OBox_egwd} and Eq-\ref{eq:OBox_edwd}, we find that the mathematical form of EGWD can be viewed as a special case of EDWD when setting identical variance $\sigma_w^2$, $\sigma_h^2$ for each side, although they are developed under different assumptions. 
For EDWD, the variance is set according to the length of the edge. We will show that this property is beneficial to the optimization process in the next part.

\noindent\textbf{Scale Normalization}. In object detection, a good regression loss should consider the problem brought by bounding box's scales. For L1 loss, the large target will produce much larger penalty than small targets, which results in the degeneration of performance of small objects. To address this issue, we normalize both the predicted and target bounding boxes by a certain scale to make the EWD distance more robust in various scales.
Specifically, we normalize the target width and height $(w_t, h_t)$ to $(1, 1)$ and use them to scale the predicted $(w_p, h_p)$ to $(\frac{w_p}{w_t}, \frac{h_p}{h_t})$. As for the center offset, using either $w_t$ or $h_t$ as the denominator is ambiguous, thus we use $\sqrt{w_t h_t}$ to scale the offset. We also use this value to scale the variance of EDWD. Then for gradient with respect to center  $\mathbf{o_p}=(x_p, y_p)$
\begin{equation}
\frac{\partial \mathbf{W_{edwd}}}{\partial \mathbf{o_p}} = \left(\frac{2}{s^2}\Delta x, \frac{2}{s^2}\Delta y \right)
\end{equation}
The scale factor $s$ will dynamically adjust the gradient for various
scaled boxes. Small targets will get large gradients which is important as that small deviation will produce much more severe IoU decrease for small scale targets. Similar analysis can be done for $w_p$ and $h_p$. 
\begin{equation}
\begin{split}
    \frac{\partial \mathbf{W_{edwd}}}{\partial w_p} &= (1 + \frac{w_t}{h_t})(\frac{w_p}{w_t} - \cos\Delta\theta), \\
    \frac{\partial f_{EWD}(h_p)}{\partial h_p} &= (1 + \frac{h_t}{w_t})(\frac{h_p}{h_t} - \cos\Delta\theta)
\end{split}
\end{equation}
We can see that the gradient of $w_p$ and $h_p$ are also adjusted by the normalization scale. Smaller height or width will receive larger gradient. Besides, the gradient is also affected by the $\theta$ difference and aspect ratio. The larger side will get slightly larger gradient compared to smaller side. This is desirable especially for extremely smaller object where regression is more likely to be affected by noise. For the gradient of $\Delta\theta$, we have
\begin{equation}\label{eq:grad_theta}
    \frac{\partial \mathbf{W_{edwd}}(\Delta\theta)}{\partial \cos\Delta\theta} = - \left(\frac{w_p}{w_t} + \frac{h_p}{h_t} + \frac{w_p}{h_t} + \frac{h_p}{w_t}\right)
\end{equation}
When $w_p = w_t$ and $h_p = h_t$, then $\frac{\partial f_{EWD}(\Delta\theta)}{\partial \cos\Delta\theta} = -(2 + \frac{w_t}{h_t} + \frac{h_t}{w_t}) \geq - 4$, the condition for equality is that $h_t = w_t$. That is the square case. So this makes the model to produce large gradient for angle when the aspect ratio is large. For the square case, there are still gradient for the angle optimization. Remind that both KLD and GWD suffers from problem of gradient vanishing in square cases. 

\noindent\textbf{Post Function}. We find that directly applying the EWD distance as the regression loss is unstable. In most cases, the regression process failed to converge. This is caused by the squared operation which is sensitive to large errors. To fix it, we simply apply a squared root operation to the EWD distance. Note that GWD and KLD also encounter this problem and they use a non-linear mapping function $\log(1+x)$ or $\frac{1}{\tau + x}$ to modulate the distance. We also tried this techniques and get similar results with the squared root operation.

\begin{figure*}[t]
\centering
\begin{subfigure}{0.23\linewidth}
    \centering
    \includegraphics[width=4cm]{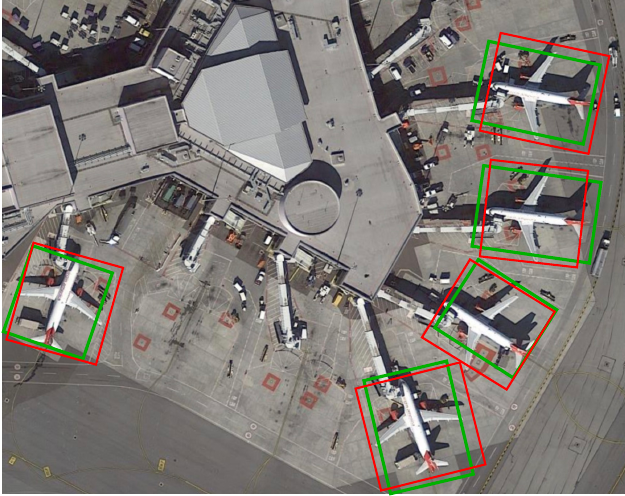}\\
    \includegraphics[width=4cm]{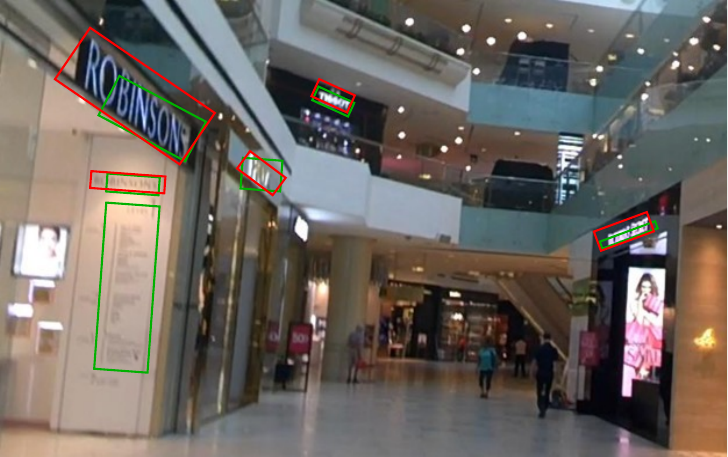}
\caption{Smooth L1}
\end{subfigure}
\begin{subfigure}{0.23\linewidth}
    \centering
    \includegraphics[width=4cm]{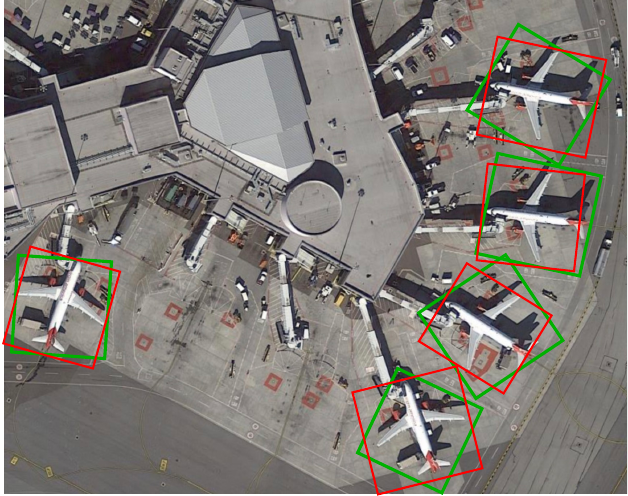}\\
    \includegraphics[width=4cm]{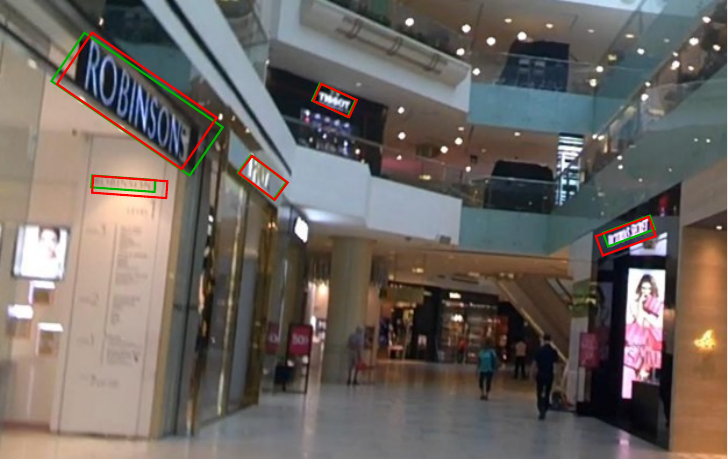}
\caption{KLD}
\end{subfigure}
\begin{subfigure}{0.23\linewidth}
    \centering
    \includegraphics[width=4cm]{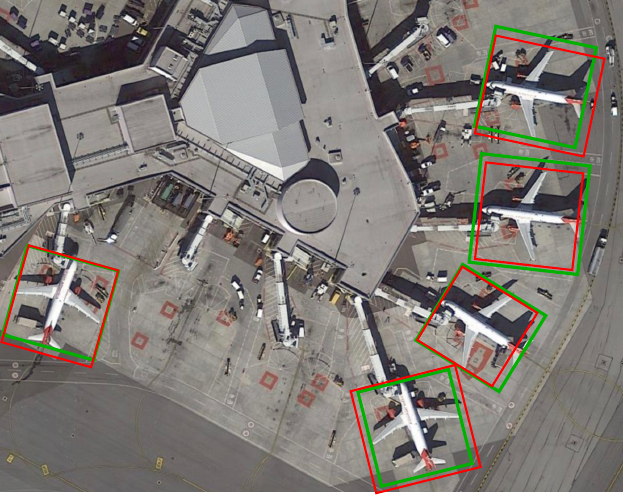}\\
    \includegraphics[width=4cm]{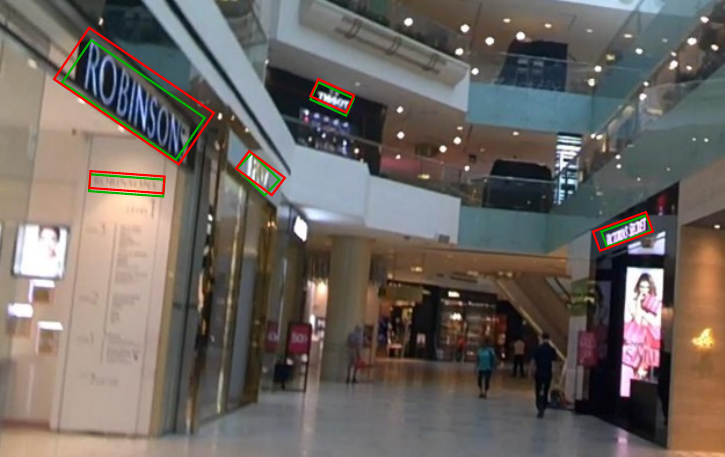}
\caption{EDWD}
\end{subfigure}
\caption{Visual comparison between different losses. Ground-truth bboxes are in \textcolor{red}{red}, predicted bboxes are in \textcolor{green}{green}.}
\label{fig:vis_compare}
\vspace{-12pt}
\end{figure*}
\begin{figure*}[t]
\centering
\begin{subfigure}{0.23\linewidth}
    \centering
    \includegraphics[width=4cm]{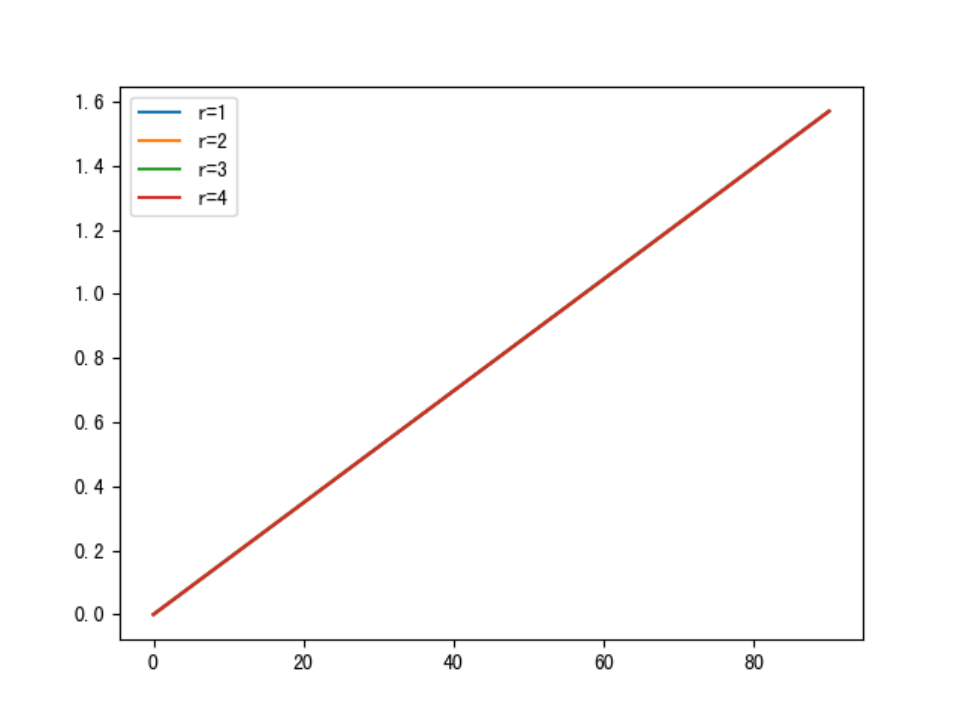}
\caption{Smooth L1}
\end{subfigure}
\begin{subfigure}{0.23\linewidth}
    \centering
    \includegraphics[width=4cm]{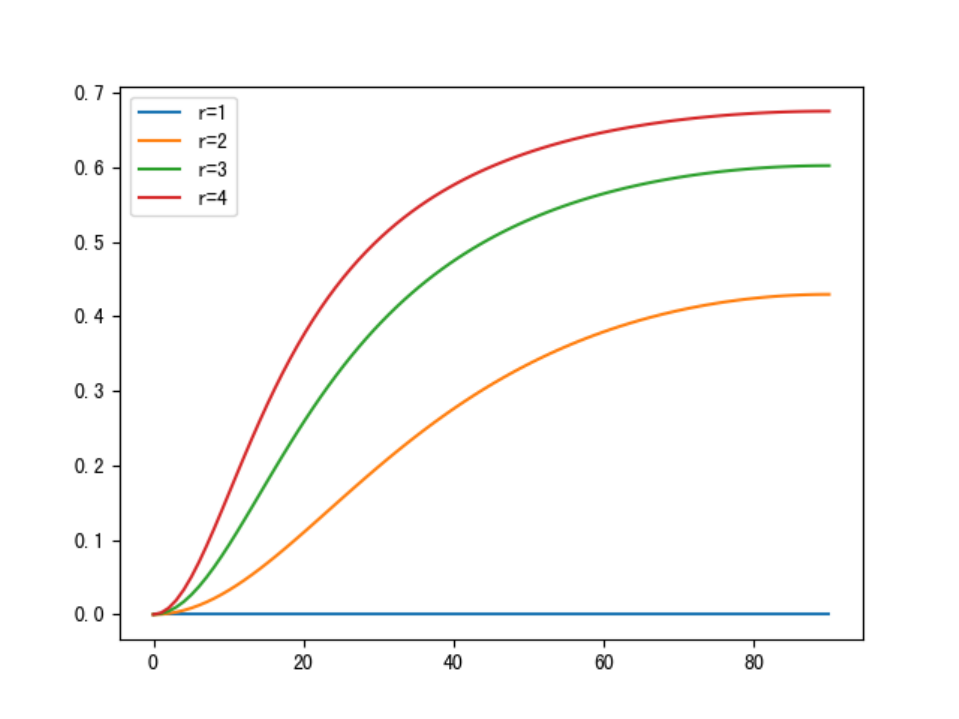}
\caption{KLD}
\end{subfigure}
\begin{subfigure}{0.23\linewidth}
    \centering
    \includegraphics[width=4cm]{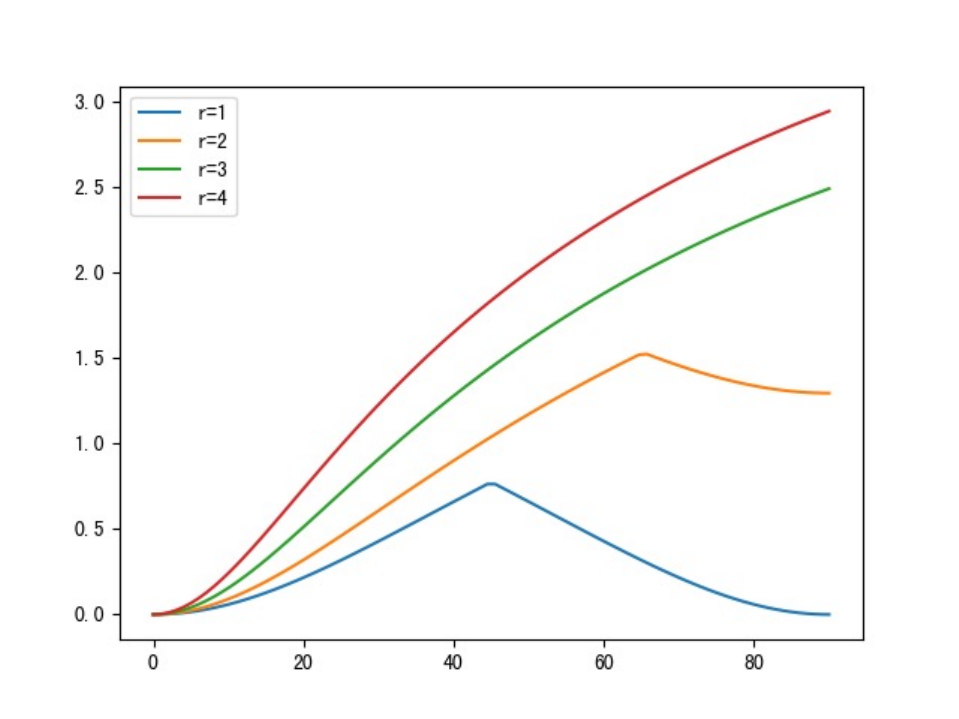}
\caption{EDWD}
\end{subfigure}
\caption{Smooth L1, KLD and EDWD versus angle when aspect ratio varies.}
\label{fig:loss_curve}
\vspace{-12pt}
\end{figure*}

\begin{table*}[h]
\centering
\begin{adjustbox}{max width=\textwidth}
\begin{tabular}{l|l|l|lllll}
\hline
\multirow{2}{*}{method}                         & \multirow{2}{*}{loss} & \multirow{2}{*}{w/o norm} & \multicolumn{5}{c}{w/ norm by}                                                 \\ \cline{4-8} 
                                                &                       &                           & image size & anchor size & $h_t$ / $w_t$ & $\min(h_t, w_t)$ & $\max(h_t, w_t)$ \\ \hline
\multicolumn{1}{c|}{\multirow{2}{*}{Retinanet}} & Smooth-L1               & 45.3                      & 63.1       & 67.2        & 67.8          & 68.2             & 68.1             \\
\multicolumn{1}{c|}{}                           & EWD                   & 67.2                      & 66.9       & 68.1        & 68.8          & \textbf{69.6}    & 69.4             \\ \hline
\end{tabular}
\end{adjustbox}
\caption{Ablations of scale normalization on DOTA-1.0. w/o and w/ are short for without and with. }
\label{tab:scale_norm}
\end{table*}

\begin{table*}[!h]
\begin{adjustbox}{max width=\textwidth}
\begin{tabular}{cccccccc|cc|ccccc}
\hline
detector  & method                                 & Box Def                   & $BR^{\dag}$                                          & $SV^{\dag}$                                                   & $LV^{\dag}$                                                   & $SH^{\dag}$                                                   & $HA^{\dag}$                                                   & $ST^{\ddag}$                                                  & $RA^{\ddag}$                                                  & 5-$AP_{50}$                                                   & 2-$AP_{50}$                                                   & $AP_{50}$                                                     & $AP_{75}$                                                     & $AP_{50:95}$                                                  \\ \hline
Retinanet & Smooth L1                              & oc                        & 34.34                                                & 55.84                                                         & 66.88                                                         & 75.16                                                         & 55.51                                                         & 83.42                                                         & 59.03                                                         & 57.54                                                         & 71.25                                                         & 63.42                                                         & 35.95                                                         & 36.07                                                         \\
          &                                        & le                        & 36.97                                                & 63.84                                                         & 70.99                                                         & 82.80                                                         & 59.33                                                         & 82.70                                                         & 64.94                                                         & 62.78                                                         & 73.82                                                         & 65.41                                                         & 36.14                                                         & 37.17                                                         \\
          &                                        & min                       & 35.21                                                & 65.18                                                         & 73.02                                                         & 85.25                                                         & 58.07                                                         & 82.23                                                         & 62.33                                                         & 63.34                                                         & 72.28                                                         & 67.26                                                         & 38.29                                                         & 39.00                                                         \\
          & IoU-Smooth L1\cite{yang2019scrdet}     & min                       & 36.44                                                & 62.25                                                         & 74.01                                                         & 86.43                                                         & 58.17                                                         & 81.08                                                         & 63.01                                                         & 63.46                                                         & 72.05                                                         & 67.19                                                         & 38.21                                                         & 39.14                                                         \\
          & Modulated Loss\cite{qian2021modulated} & -                         & 36.30                                                & 64.01                                                         & 69.33                                                         & 82.76                                                         & 61.20                                                         & 82.53                                                         & 61.29                                                         & 62.72                                                         & 71.91                                                         & 66.31                                                         & 37.32                                                         & 38.61                                                         \\
          & GWD\cite{yang2021rethinking}           & -                         & {\color[HTML]{FE0000} \textbf{41.03}}                & 66.30                                                         & 73.80                                                         & 83.31                                                         & 61.76                                                         & 83.82                                                         & 63.33                                                         & 65.24                                                         & 73.57                                                         & 68.61                                                         & 40.68                                                         & 40.71                                                         \\
          & KLD\cite{yang2021learning}             & -                         & {\color[HTML]{3531FF} \textbf{40.97}}                & {\color[HTML]{FE0000} \textbf{67.21}}                         & {\color[HTML]{FE0000} \textbf{76.52}}                         & {\color[HTML]{3531FF} \textbf{86.20}}                         & {\color[HTML]{FE0000} \textbf{63.10}}                         & {\color[HTML]{3531FF} \textbf{84.08}}                         & 66.01                                                         & {\color[HTML]{FE0000} \textbf{66.8}}                          & 75.04                                                         & 68.14                                                         & {\color[HTML]{FE0000} \textbf{44.48}}                         & {\color[HTML]{FE0000} \textbf{42.15}}                         \\
          & \cellcolor[HTML]{EFEFEF}EGWD           & \cellcolor[HTML]{EFEFEF}- & \cellcolor[HTML]{EFEFEF}36.26                        & \cellcolor[HTML]{EFEFEF}66.16                                 & \cellcolor[HTML]{EFEFEF}72.15                                 & \cellcolor[HTML]{EFEFEF}85.54                                 & \cellcolor[HTML]{EFEFEF}61.20                                 & \cellcolor[HTML]{EFEFEF}84.02                                 & \cellcolor[HTML]{EFEFEF}{\color[HTML]{3531FF} \textbf{66.20}} & \cellcolor[HTML]{EFEFEF}64.26                                 & \cellcolor[HTML]{EFEFEF}{\color[HTML]{3531FF} \textbf{75.11}} & \cellcolor[HTML]{EFEFEF}{\color[HTML]{3531FF} \textbf{69.10}} & \cellcolor[HTML]{EFEFEF}42.62                                 & \cellcolor[HTML]{EFEFEF}40.73                                 \\
          & \cellcolor[HTML]{EFEFEF}EDWD           & \cellcolor[HTML]{EFEFEF}- & \cellcolor[HTML]{EFEFEF}39.57                        & \cellcolor[HTML]{EFEFEF}{\color[HTML]{3531FF} \textbf{66.95}} & \cellcolor[HTML]{EFEFEF}{\color[HTML]{3531FF} \textbf{76.29}} & \cellcolor[HTML]{EFEFEF}{\color[HTML]{FE0000} \textbf{87.01}} & \cellcolor[HTML]{EFEFEF}{\color[HTML]{3531FF} \textbf{62.01}} & \cellcolor[HTML]{EFEFEF}{\color[HTML]{FE0000} \textbf{84.25}} & \cellcolor[HTML]{EFEFEF}{\color[HTML]{FE0000} \textbf{66.45}} & \cellcolor[HTML]{EFEFEF}{\color[HTML]{3531FF} \textbf{66.36}} & \cellcolor[HTML]{EFEFEF}{\color[HTML]{FE0000} \textbf{75.35}} & \cellcolor[HTML]{EFEFEF}{\color[HTML]{FE0000} \textbf{69.61}} & \cellcolor[HTML]{EFEFEF}{\color[HTML]{3531FF} \textbf{43.60}} & \cellcolor[HTML]{EFEFEF}{\color[HTML]{3531FF} \textbf{41.66}} \\ \hline
FCOS      & Smooth L1-Loss                         & min                       & 39.60                                                & 67.70                                                         & 72.91                                                         & 85.04                                                         & 60.65                                                         & 83.03                                                         & 67.40                                                         & 65.18                                                         & 75.22                                                         & 68.42                                                         & 29.46                                                         & 35.30                                                         \\
          & GWD\cite{yang2021rethinking}           & -                         & {\color[HTML]{3531FF} \textbf{42.73}}                & 68.12                                                         & {\color[HTML]{3531FF} \textbf{82.48}}                         & {\color[HTML]{3531FF} \textbf{87.86}}                         & 65.90                                                         & 83.67                                                         & 67.31                                                         & 69.62                                                         & 75.39                                                         & {\color[HTML]{3531FF} \textbf{70.00}}                         & 38.20                                                         & 39.70                                                         \\
          & KLD\cite{yang2021learning}             & -                         & {\color[HTML]{FE0000} \textbf{43.52}}                & {\color[HTML]{FE0000} \textbf{69.70}}                         & 81.67                                                         & 87.41                                                         & 65.48                                                         & 84.82                                                         & {\color[HTML]{3531FF} \textbf{67.87}}                         & {\color[HTML]{3531FF} \textbf{69.56}}                         & 75.35                                                         & 69.94                                                         & 39.52                                                         & {\color[HTML]{3166FF} \textbf{40.93}}                         \\
          & \cellcolor[HTML]{EFEFEF}EGWD           & \cellcolor[HTML]{EFEFEF}- & \cellcolor[HTML]{EFEFEF}41.79                        & \cellcolor[HTML]{EFEFEF}68.12                                 & \cellcolor[HTML]{EFEFEF}81.79                                 & \cellcolor[HTML]{EFEFEF}87.21                                 & \cellcolor[HTML]{EFEFEF}{\color[HTML]{3531FF} \textbf{67.31}} & \cellcolor[HTML]{EFEFEF}{\color[HTML]{3531FF} \textbf{84.95}} & \cellcolor[HTML]{EFEFEF}67.80                                 & \cellcolor[HTML]{EFEFEF}69.24                                 & \cellcolor[HTML]{EFEFEF}{\color[HTML]{3531FF} \textbf{76.38}} & \cellcolor[HTML]{EFEFEF}69.83                                 & \cellcolor[HTML]{EFEFEF}{\color[HTML]{3531FF} \textbf{41.38}} & \cellcolor[HTML]{EFEFEF}40.26                                 \\
          & \cellcolor[HTML]{EFEFEF}EDWD           & \cellcolor[HTML]{EFEFEF}- & \cellcolor[HTML]{EFEFEF}42.01                        & \cellcolor[HTML]{EFEFEF}{\color[HTML]{3531FF} \textbf{68.52}} & \cellcolor[HTML]{EFEFEF}{\color[HTML]{FE0000} \textbf{82.78}} & \cellcolor[HTML]{EFEFEF}{\color[HTML]{FE0000} \textbf{88.00}} & \cellcolor[HTML]{EFEFEF}{\color[HTML]{FE0000} \textbf{70.73}} & \cellcolor[HTML]{EFEFEF}{\color[HTML]{FE0000} \textbf{85.41}} & \cellcolor[HTML]{EFEFEF}{\color[HTML]{FE0000} \textbf{68.82}} & \cellcolor[HTML]{EFEFEF}{\color[HTML]{FE0000} \textbf{70.41}} & \cellcolor[HTML]{EFEFEF}{\color[HTML]{FE0000} \textbf{77.12}} & \cellcolor[HTML]{EFEFEF}{\color[HTML]{FE0000} \textbf{70.91}} & \cellcolor[HTML]{EFEFEF}{\color[HTML]{FE0000} \textbf{41.45}} & \cellcolor[HTML]{EFEFEF}{\color[HTML]{FE0000} \textbf{41.51}} \\ \hline
S2ANet    & Smooth L1-Loss                         & -                         & 44.68                                                & 66.17                                                         & 81.15                                                         & 88.00                                                         & 68.84                                                         & 85.54                                                         & 65.17                                                         & 69.97                                                         & 75.35                                                         & 70.05                                                         & 40.71                                                         & 39.80                                                         \\
          & \cellcolor[HTML]{EFEFEF}EDWD           & \cellcolor[HTML]{EFEFEF}- & \cellcolor[HTML]{EFEFEF}{\color[HTML]{FE0000} 45.01} & \cellcolor[HTML]{EFEFEF}{\color[HTML]{FE0000} \textbf{67.14}} & \cellcolor[HTML]{EFEFEF}{\color[HTML]{FE0000} \textbf{81.60}} & \cellcolor[HTML]{EFEFEF}{\color[HTML]{FE0000} \textbf{88.32}} & \cellcolor[HTML]{EFEFEF}{\color[HTML]{FE0000} \textbf{68.90}} & \cellcolor[HTML]{EFEFEF}{\color[HTML]{FE0000} \textbf{85.90}} & \cellcolor[HTML]{EFEFEF}{\color[HTML]{FE0000} \textbf{65.30}} & \cellcolor[HTML]{EFEFEF}{\color[HTML]{FE0000} \textbf{70.19}} & \cellcolor[HTML]{EFEFEF}{\color[HTML]{FE0000} \textbf{75.60}} & \cellcolor[HTML]{EFEFEF}{\color[HTML]{FE0000} \textbf{71.08}} & \cellcolor[HTML]{EFEFEF}{\color[HTML]{FE0000} \textbf{41.50}} & \cellcolor[HTML]{EFEFEF}{\color[HTML]{FE0000} \textbf{40.03}} \\ \hline
O-RCNN    & Smooth L1-Loss                         & -                         & {\color[HTML]{FE0000} \textbf{48.46}}                & 66.17                                                         & 84.51                                                         & 88.33                                                         & {\color[HTML]{FE0000} \textbf{75.97}}                         & {\color[HTML]{FE0000} \textbf{80.88}}                         & 70.15                                                         & {\color[HTML]{FE0000} \textbf{72.69}}                         & 75.51                                                         & 73.56                                                         & 47.78                                                         & 44.55                                                         \\
          & \cellcolor[HTML]{EFEFEF}EDWD           & \cellcolor[HTML]{EFEFEF}- & \cellcolor[HTML]{EFEFEF}48.43                        & \cellcolor[HTML]{EFEFEF}{\color[HTML]{FE0000} \textbf{66.18}} & \cellcolor[HTML]{EFEFEF}{\color[HTML]{FE0000} \textbf{84.58}} & \cellcolor[HTML]{EFEFEF}{\color[HTML]{FE0000} \textbf{88.40}} & \cellcolor[HTML]{EFEFEF}75.80                                 & \cellcolor[HTML]{EFEFEF}80.72                                 & \cellcolor[HTML]{EFEFEF}{\color[HTML]{FE0000} \textbf{70.31}} & \cellcolor[HTML]{EFEFEF}72.68                                 & \cellcolor[HTML]{EFEFEF}{\color[HTML]{FE0000} \textbf{75.52}} & \cellcolor[HTML]{EFEFEF}{\color[HTML]{FE0000} \textbf{73.56}} & \cellcolor[HTML]{EFEFEF}{\color[HTML]{FE0000} \textbf{47.80}} & \cellcolor[HTML]{EFEFEF}{\color[HTML]{FE0000} \textbf{44.56}} \\ \hline
\end{tabular}
\end{adjustbox}
\caption{Comparisons on DOTA-1.0. Refer to Sec-\ref{sec:datasets} for short names. The symbols, $\dag$, $\circ$, $\square$, represent large aspect ratio, circular and near-square objects respectively. The bold \textbf{\textcolor{red} {red}} and \textbf{\textcolor{blue} {blue}} fonts indicate the top two performance. The oc, le and min denote the opencv definition($\theta \in [-90^{\circ}, 0^{\circ})$) , long edge definition and minimum theta definition($\theta \in [-45^{\circ}, 45^{\circ})$ of oriented bounding box.
}
\label{tab:dota}
\vspace{-12pt}
\end{table*}

\section{Experiments}
\subsection{Datasets and implementation details}
\label{sec:datasets}
We evaluate our method on multiple datasets mainly covering aerial images: DOTA\cite{xia2018dota}, HRSC2016\cite{liu2017hrsc}, and text detection: ICDAR2015\cite{karatzas2015icdar}, MSRA-TD500\cite{yao2012msra_tda}.

\textbf{DOTA}\cite{xia2018dota} is one of the largest dataset for multi-class object detection in aerial images. It contains 15 categories, 2,806 images and 188,282 instances. Images' scales range from $800 \times 800$ to $4000 \times 4000$ pixels. In the dataset, each object is annotated by an oriented bounding box (OBox), which can be denoted as ($x_1,y_1,x_2,y_2,x_3,y_3,x_4,y_4$) , where ($x_i,y_i$) denotes the i-th vertice of OBox. The dataset is split into training, validation, and test sets with 1/2, 1/6, and 1/3 ratio, respectively. For the short names in Table-\ref{tab:dota}, they are defined as(abbreviation-full name): BR-Bridge, SV-Small vehicle, LV-Large vehicle, SH-Ship, HA-Harbor, ST-Storage tank, RA-Roundabout, PL-Plane, BD-Baseball diamond.

\textbf{HRSC2016}\cite{liu2017hrsc} is another aerial images data set for ship detection. It contains images from two scenarios including ships on sea and ships close inshore. HRSC2016 dataset has four ship categories, 1061 images and 2976 samples in total. It split into training, validation and test set which contains 436 images including 1207 samples, 181 images including 541 samples and 444 images respectively. The image sizes range from $300 \times 300$, to $1500 \times 900$.

\textbf{ICDAR2015}\cite{karatzas2015icdar} and \textbf{MSRA-TD500}\cite{yao2012msra_tda} are both commonly used for oriented scene text detection and spotting. ICDAR2015 includes 1,000 training images and 500 testing images. MSRA-TD500 consists of 300 training images and 200 testing images. We perform experiments on these scene text detection datasets in order to verify the scenario generality of our work.

We mainly choose two types of detectors, Retinanet\cite{2017Focal} and FCOS\cite{2020FCOS} to verify our methods. The implementation of the rotated version of them are based on Alpharotate\cite{yang2021alpharotate} and MMRotate\cite{mmrotate2022}. These two detectors are specially chosen as the representation of anchor-based methods and anchor free methods. 
We use ResNet-50\cite{he2016resnet} backbone with FPN\cite{2017fpn} head for all datasets.
Training is conducted on single Tesla P100 GPU with 16GB memory, each with batch size of 2. For each datasets, Stochastic gradient descent(SGD) is adopted as the optimizer and the base learning rate is set to 0.0025, weight decay set to 0.0001, momentum set to 0.9. 
The training details for each dataset is somewhat different. For DOTA, we use the training split to train the detector 20 epochs and use the validation split to report the final results. For HRSC, ICDAR2015 and MSRA-TD500, as the these datasets are relatively small, we set the training step of one epoch to 2190, 5000 and 1500 and total epochs to 12 to ensure that the detector is fully trained. The base augmentation is simply set as random flip. If other augmentation, eg, random rotate, is applied, the training length will be increased accordingly.

\subsection{Comparisons and Ablations}

\textbf{Post functions}. 
Table-\ref{tab:post} compares the performance of different post functions and hyper parameters. We report the results on HRSC2016 with Retinanet. Other datasets and detectors will result in similar conclusions. The results show that the EWD without post function gets poor performance. Thus, a simple $\sqrt x$ function will improve the performance by a large margin. Using the $\log(1+x)$ as the post function gets the best performance $89.49\%$. For the post function $1 - \frac{1}{\tau+x}$ proposed by GWD\cite{yang2021rethinking}, we find that the hyper parameter $\tau$ hardly affects the performance. In most of the experiments, we use $\log(1+x)$ as the post function unless explicitly specified.

\textbf{Scale normalization}. Table-\ref{tab:scale_norm} shows the results with different types of scale normalization. The results are reported on DOTA-1.0 with Retinanet. We also investigate into the effect of scale normalization on smooth L1 loss. The results show that the smooth L1 loss without any normalization gets the worst results, while a simple normalization by either the image size or the anchor size will largely boost the performance(from 45.3 to 63.1 or 67.2). In practice, we find that the L1 loss without any normalization will get a large loss value, which will overwhelm the classification loss and thus deteriorate the training. So normalization by the image size or the anchor size will fix this problem by means of lowering down the influence of regression loss. 
Further more, by adopting the scale normalization by the target box value $h_t/w_t$ or the minimum/maximum of them, the performance can still be improved. Normalizing by the target minimum or maximum gets the best results, both for smooth L1 loss and EWD loss. In all cases, the proposed EWD loss beats the smooth l1 loss. Even without normalization, EWD loss gets 67.2\% thanks to the modulation effect of post functions. 

\textbf{Bounding Box Definition}. The bounding box definition refers to the range restriction techniques for L1 loss(referring to Sec-\ref{sec:box_rep}). In Table-\ref{tab:dota}, we show the results of smooth L1 loss with different bounding box definitions. We find that the minimum angle definition(restricting the angle to be in $[-45^{\circ}, 45^{\circ})$ shows large advantage over the other two definitions. However, in other two scene text datasets, we find that the long edge definition usually works best, which is chosen as the baseline.

\textbf{Square-like problem}. In Table-\ref{tab:dota}, we compare the performance on circular targets:storage tank(ST), roundabout(RA) and square-like targets:plane(PL), baseball diamond(BD).
The circular target is a special kind of square-like target whose angular direction is not defined. 
It is shown that EWD loss gets slightly higher results than KLD and GWD for circular objects. They all behave much better than the smooth L1 loss and its variants. 
For square-like objects, we report the $AP_{85}$ on PL and BD. The $AP_{50}$ is not chosen because of squares with only angle difference will always get IoU higher than 82\%. It is shown that the KLD and GWD methods behaves even worse than Smooth L1 loss while our method performs the best.
Moreover, we visualize the OBox predicted by L1 loss, KLD methods and ours. It is shown in Figure-\ref{fig:vis_compare}. For KLD loss, it fails to predict the direction of the square object. But the L1 loss and ours do not suffer from it. 
The square-like problem can also be understood by the distance-angle curve shown in Figure-\ref{fig:loss_curve}. 
We plot the distance compared by L1-loss, KLD and EWD on different aspect ratio objects. For square-like objects(aspect r=1, the blue curve), the distance of KLD is always zero. While L1 and EWD do not have this problem.

\begin{table}[!h]
\centering
\resizebox{0.46 \textwidth}{!}{
\begin{tabular}{c|l|llll}
\hline
\multicolumn{1}{l|}{\multirow{2}{*}{}} & \multirow{2}{*}{$f(W)$} & \multicolumn{4}{c}{$1 - \frac{1}{\tau + f(W)}$} \\ \cline{3-6} 
\multicolumn{1}{l|}{}                  &                         & $\tau=1$   & $\tau=2$   & $\tau=3$  & $\tau=5$  \\ \hline
$f(W) = W$                             & 60.01                   & 88.78      & 88.91      & 88.66     & 88.51     \\
$f(W) = \sqrt{W}$                      & 88.27                   & 88.45      & 87.90      & 88.21     & 88.35     \\
$f(W) = \log(1+W)$                     & \bf{89.49}              & 89.21      & 88.13      & 87.72     & 88.21     \\ \hline
\end{tabular}
}
\caption{Ablations on post functions. The results are $AP_{50}$ reported on HRSC2016\cite{liu2017hrsc}. }
\label{tab:post}
\end{table}

\textbf{Comparison with peer methods}.
In this part, we compare EWD loss with smooth L1 loss, IoU-Smooth L1 Loss\cite{yang2019scrdet}, Modulated  loss\cite{qian2021modulated}, GWD\cite{yang2021rethinking} loss and KLD\cite{yang2021learning} loss. These methods are chosen as they are all designed to solve the problems for oriented regression. GWD and KLD are most relevant to our method and the comparison with them is conducted on all datasets and settings.
The comparison is conducted on three commonly used datasets, DOTA-1.0(Table-\ref{tab:dota}), HRSC2016(Table-\ref{tab:hrsc}),  ICDAR2015(Table-\ref{tab:icdar}) and MSRA-TD500(Table-\ref{tab:msra}). We choose two most representative detectors, Retinanet and FCOS, and implement their rotated version as the base detectors. 
In DOTA dataset, the results with two state-of-the-art detectors, S2ANet\cite{S2ANet} and Oriented RCNN\cite{orcnn} are reported.
We can see that in S2ANet and Oriented RCNN, where the baseline is very high, the performane can still be improved.
The rotated FCOS increases overall baseline performance compared to rotated Retinanet, but it gets a worse result on high-IoU occasions. Nevertheless, we can see that on all occasions, the EDWD loss gets the best results. For example, using rotated Retinanet on DOTA-1.0, the EDWD loss increased the $AP_{50}$ by 2.35\% compared to the best performed Smooth L1 loss. It also beats the KLD loss by 1.47\%. 

\begin{table}[t]
\centering
\resizebox{0.46 \textwidth}{!}{
\begin{tabular}{l|l|lllll}
\hline
Detector  & Methods & Smooth L1 & \begin{tabular}[c]{@{}l@{}}GWD\\ \end{tabular} & \begin{tabular}[c]{@{}l@{}}KLD\\ \end{tabular} & EGWD  & EDWD           \\ \hline
Retinanet & AP50    & 84.28     & 85.56                                                                   & 87.45                                                                 & 88.87 & \textbf{89.49} \\
          & AP50:95 & 47.76     & 52.89                                                                   & 58.72                                                                 & 56.19 & \textbf{58.80} \\ \hline
FCOS      & AP50    & 88.30     & 89.74                                                                   & 89.85                                                                 & 89.70 & \textbf{90.30} \\
          & AP50:95 & 51.82     & 54.12                                                                   & 55.03                                                                 & 53.98 & \textbf{55.32} \\ \hline
\end{tabular}
}
\caption{Experiments on HRSC\cite{liu2017hrsc}.}
\label{tab:hrsc}
\vspace{-6pt}
\end{table}

\begin{table*}[ht]
\centering
\begin{adjustbox}{max width=\textwidth}
\begin{tabular}{cc|cccccc}
\hline
                            &                              & \multicolumn{3}{c}{No Rotate Aug}                                                                                        & \multicolumn{3}{c}{Rotate Aug}                                                                                           \\ \cline{3-8} 
\multirow{-2}{*}{Detector}  & \multirow{-2}{*}{Method}     & Hmean50                                & Hmean75                                & Hmean50:95                             & Hmean50                                & Hmean75                                & Hmean50:95                             \\ \hline
                            & Smooth L1-Loss               & 71.52                                  & 45.62                                  & 41.98                                  & 74.18                                  & 48.39                                  & 43.39                                  \\
                            & GWD                          & 74.18                                  & 46.22                                  & 43.23                                  & 75.82                                  & 49.64                                  & 45.53                                  \\
                            & KLD                          & 74.41                                  & 47.05                                  & 43.12                                  & 76.41                                  & 50.41                                  & 45.30                                  \\
                            & \cellcolor[HTML]{EFEFEF}EGWD & \cellcolor[HTML]{EFEFEF}74.62          & \cellcolor[HTML]{EFEFEF}46.91          & \cellcolor[HTML]{EFEFEF}43.45          & \cellcolor[HTML]{EFEFEF}76.86          & \cellcolor[HTML]{EFEFEF}50.92          & \cellcolor[HTML]{EFEFEF}45.69          \\
\multirow{-5}{*}{Retinanet} & \cellcolor[HTML]{EFEFEF}EDWD & \cellcolor[HTML]{EFEFEF}\textbf{74.77} & \cellcolor[HTML]{EFEFEF}\textbf{47.60} & \cellcolor[HTML]{EFEFEF}\textbf{43.73} & \cellcolor[HTML]{EFEFEF}\textbf{77.49} & \cellcolor[HTML]{EFEFEF}\textbf{51.23} & \cellcolor[HTML]{EFEFEF}\textbf{46.18} \\ \hline
                            & L1                           & 70.71                                  & 43.01                                  & 41.15                                  & 74.01                                  & 47.64                                  & 43.27                                  \\
                            & GWD                          & 74.50                                  & 44.27                                  & 42.19                                  & 75.78                                  & 48.52                                  & 44.61                                  \\
                            & KLD                          & 74.71                                  & \textbf{47.37}                                & 42.25                                  & 76.12                                  & \textbf{49.31}                                & 44.53                                  \\
                            & \cellcolor[HTML]{EFEFEF}EGWD & \cellcolor[HTML]{EFEFEF}74.31          & \cellcolor[HTML]{EFEFEF}45.89          & \cellcolor[HTML]{EFEFEF}43.72          & \cellcolor[HTML]{EFEFEF}75.65          & \cellcolor[HTML]{EFEFEF}47.99          & \cellcolor[HTML]{EFEFEF}44.70          \\
\multirow{-5}{*}{FCOS}      & \cellcolor[HTML]{EFEFEF}EDWD & \cellcolor[HTML]{EFEFEF}\textbf{74.94} & \cellcolor[HTML]{EFEFEF}46.29 & \cellcolor[HTML]{EFEFEF}\textbf{43.14} & \cellcolor[HTML]{EFEFEF}\textbf{76.33} & \cellcolor[HTML]{EFEFEF}49.28 & \cellcolor[HTML]{EFEFEF}\textbf{45.07} \\ \hline
\end{tabular}
\end{adjustbox}
\caption{Comparisons on ICDAR2015}
\label{tab:icdar}
\vspace{-6pt}
\end{table*}

\begin{table*}[!h]
\centering
\begin{adjustbox}{max width=\textwidth}
\begin{tabular}{cc|cccccc}
\hline
                            &                              & \multicolumn{3}{c}{No Rotate Aug}                                                                                        & \multicolumn{3}{c}{Rotate Aug}                                                                                           \\ \cline{3-8} 
\multirow{-2}{*}{Detector}  & \multirow{-2}{*}{Method}     & Hmean50                                & Hmean75                                & Hmean50:95                             & Hmean50                                & Hmean75                                & Hmean50:95                             \\ \hline
                            & Smooth L1-Loss               & 64.34                                  & 38.86                                  & 38.39                                  & 70.24                                  & 41.94                                  & 40.20                                  \\
                            & GWD                          & 66.47                                  & 43.61                                  & 39.96                                  & 72.66                                  & 43.72                                  & 42.31                                  \\
                            & KLD                          & 66.21                                  & 46.86                                  & 41.94                                  & 73.55                                  & 45.44                                  & 42.14                                  \\
                            & \cellcolor[HTML]{EFEFEF}EGWD & \cellcolor[HTML]{EFEFEF}67.10          & \cellcolor[HTML]{EFEFEF}47.32          & \cellcolor[HTML]{EFEFEF}42.01          & \cellcolor[HTML]{EFEFEF}73.98          & \cellcolor[HTML]{EFEFEF}48.64          & \cellcolor[HTML]{EFEFEF}44.10          \\
\multirow{-5}{*}{Retinanet} & \cellcolor[HTML]{EFEFEF}EDWD & \cellcolor[HTML]{EFEFEF}\textbf{67.32} & \cellcolor[HTML]{EFEFEF}\textbf{48.65} & \cellcolor[HTML]{EFEFEF}\textbf{42.55} & \cellcolor[HTML]{EFEFEF}\textbf{74.13} & \cellcolor[HTML]{EFEFEF}\textbf{49.14} & \cellcolor[HTML]{EFEFEF}\textbf{45.08} \\ \hline
\end{tabular}
\end{adjustbox}
\caption{Comparisons on MSRA-TD500}
\label{tab:msra}
\end{table*}

\begin{table}[!h]
\centering
\resizebox{0.46 \textwidth}{!}{
\begin{tabular}{l|llll}
\hline
method   & Smooth-L1 & Modulated Loss\cite{qian2021modulated} & Gliding Vertex\cite{xu2020gliding} & EDWD  \\ \hline
AP50     & 72.06     & 88.51          & 87.23         & \bf{90.01} \\
mAP50:95 & 35.13     & 53.07          & 50.33         & \bf{59.62} \\ \hline
\end{tabular}
}
\caption{Results of quadrilateral regression on HRSC2016. }
\label{tab:poly}
\end{table}

\textbf{Large aspect ratio optimization}. 
The EWD loss and KLD are both optimized for large aspect ratio targets. Mathematically, for fixed angle deviation, larger aspect ratio targets result in  much larger penalty, as shown in Figure-\ref{fig:loss_curve}. While it makes no difference for L1 loss. 
It is noticed that there is a sharp turning point on the loss curve of EWD loss for small aspect ratio objects. As long as the aspect ratio is larger than a certain limit, the turn point vanishes. The turning point is caused by the infimum operation in Eq-\ref{eq:wass_final}. It means that when the angle difference is larger than the boundary value, the current bipartite match has to change in order to get the minimum. 
In Table-\ref{tab:dota}, we choose several categories that have large aspect ratio to report the performance. We use 5-$AP_{50}$ to denote their average. The scene text datasets are also large aspect ratio dominated. It is shown that EWD loss gets comparable results with KLD loss on these targets. In most times, the EWD loss exhibits better performance compared with KLD loss.

We notice that on the high IoU occasions, KLD sometimes gets higher results than EWD loss, eg, on DOTA-1.0, KLD gets 44.48\%, higher than 43.60\% of EDWD loss. It is due to that the KLD loss is designed to be highly sensitive to the aspect ratio change. We refer the readers to compare the gradient expression of Equation-15 on KLD and Equ-\ref{eq:grad_theta}. Anyway, the overall performance of EWD loss surpass all the peer methods.

\textbf{Polynomial Regression} To verify the performance on general polynomial regressions, we design a polynomial regression head based on FCOS\cite{2020FCOS}. The regression branch regresses four offset vectors $(\delta x, \delta y)$ relative to the anchor point. The other part of the detector is kept identical to the original FCOS. For comparison, we choose smooth L1 loss, modulated loss\cite{qian2021modulated}, gliding vertex\cite{xu2020gliding}. These methods are chosen for they can also be applied to the quadrilateral regression occasion. For smooth L1 loss, the regression target is sorted in the order of top point, right point, bottom point, and left point. Note that gliding vertex regresses a set of specially designed parameters, horizontal width/height, four gliding offset(referring to the original paper for details). We follow their setting and modify the regression head correspondingly. The comparison results are shown in Table-\ref{tab:poly}. From the results, we find that for L1 loss, although the regression targets are sorted in order, the optimization is still unstable and tend to converge slowly. The modulated loss gets the second best results and the proposed EWD loss gets the best, even higher mAP(59.62vs55.32) than in the oriented case.

\section{Conclusion}
In this paper, we propose a novel distance measure based on the Wasserstein distance of bounding box's edges. By making some assumptions, we develop a novel loss function for oriented object detection, aka, Edge Wasserstein Distance loss (EWD). It is applicable to the general polynomial regression case, including the commonly seen oriented bonding box regression and quadrilateral regression. Interestingly, we find that for the oriented bounding box, the EWD loss turns out to be a generalized form of L2-distance loss. To verify the proposed method, extensive experiments are conducted on aerial images and scene text images. It demonstrates that the proposed method gets satisfactory results. 

{\small
\bibliographystyle{ieee_fullname}
\bibliography{egbib}
}

\end{document}


\title{Edge Wasserstein Distance Loss for Oriented Object Detection}  

\maketitle
\thispagestyle{empty}
\appendix

\section{Appendix}

\subsection{Further Comparison on DOTA}
We find that longer training will further improve the performance on DOTA dataset. Usually, when training for 36 epochs, the performance starts to saturate. 
For example, when training for 36 epochs, the retinanet gets 70.3 in AP50 metric. It increases by 3\% compared to 12-epoch training.
the S2ANet gets 73.0 in AP50, while 12-epoch training only gets 70.0.
So in this part, we make a full comparison for the commonly used loss functions in such situation. 
The results is shown in Table-\ref{tab:long_training_compare}.
We can see that when training for longer, the performance gap between losses are almost vanished. The KLD and GWD even gets inferior mAP performance than the baseline L1 loss. While the proposed EWD loss can still improve the performance under such long training protocol. It validates the performance of the proposed EWD loss.
\begin{table}[H]
\begin{adjustbox}{max width=\linewidth}
\begin{tabular}{c|c|c|ccc}
\hline
Detector  & Method & Epochs & AP50          & AP75          & AP50:95       \\ \hline
Retinanet & L1     & 36     & 70.3          & 45.5          & 43.5          \\
          & GWD    & 36     & 70.4          & 42.7          & 42.0          \\
          & KLD    & 36     & 70.5          & 44.8          & 43.3          \\
          & EDWD   & 36     & \textbf{70.5} & \textbf{46.5} & \textbf{44.0} \\ \hline
S2ANet    & L1     & 36     & 73.0          & 47.1          & 44.8          \\
          & GWD    & 36     & 73.5          & 45.1          & 43.6          \\
          & KLD    & 36     & \textbf{73.6} & 46.7          & 44.4          \\
          & EDWD   & 36     & 73.5          & \textbf{48.5} & \textbf{45.5} \\ \hline
\end{tabular}
\end{adjustbox}
\caption{Results under longer training epochs.}
\label{tab:long_training_compare}
\end{table}

\subsection{Full Derivation of Edge Gaussian Wasserstein Distance}
The Wasserstein distance between two Gaussian distributions is written as:
\begin{equation}\label{eq:wass_gaussian}
    \bf{W_{12}} = ||\mu_1 - \mu_2||_2^2 + \textbf{Tr}\left(\Sigma_1 + \Sigma_2 - 2(\Sigma_1^{1/2}\Sigma_2\Sigma_1^{1/2})^{1/2}\right)
\end{equation}
where $\mu_i$ is the edge center point and $\Sigma_i$ is the covariance matrix and is defined as
\begin{equation}
\begin{aligned}
&\mathbf{\Sigma_i} =\mathbf{R_i S_i^2 R_i}^{\top} \\
&=\left(\begin{array}{cc}
\cos \theta_i & -\sin \theta_i \\
\sin \theta_i & \cos \theta_i
\end{array}\right)\left(\begin{array}{cc}
\frac{w_i^2}{4} & 0 \\
0 & 0
\end{array}\right)\left(\begin{array}{cc}
\cos \theta_i & \sin \theta_i \\
-\sin \theta_i & \cos \theta_i
\end{array}\right) \\
&=\left(\begin{array}{cc}
\frac{w_i^2}{4} \cos ^{2} \theta_i & \frac{w_i^2}{4} \cos \theta_i \sin \theta_i \\
\frac{w_i^2}{4} \cos \theta_i \sin \theta_i & \frac{w_i^2}{4} \sin ^{2} \theta_i
\end{array}\right) \\
\end{aligned}
\end{equation}
where $w$ represents the edge's length and $\theta_i$ represents the angle of the edge.
Then, for the latter form of Eq-\ref{eq:wass_gaussian}, it can be simplified by
\begin{equation}
    \begin{split}
        \textbf{Tr}(...) &= \textbf{Tr}(\mathbf{\Sigma_1}) + \textbf{Tr}(\mathbf{\Sigma_2}) - 2 * \textbf{Tr}((\Sigma_1^{1/2}\Sigma_2\Sigma_1^{1/2})^{1/2}) \\
        &= \frac{w_1^2}{4} + \frac{w_2^2}{4} - 2 * \textbf{Tr}((\Sigma_1^{1/2}\Sigma_2\Sigma_1^{1/2})^{1/2})
    \end{split}
\end{equation}
Then, for the term $((\Sigma_1^{1/2}\Sigma_2\Sigma_1^{1/2})^{1/2}$, note that $\Sigma_i^{1/2} = R_iS_iR_i^{\top}$, we have
\begin{equation}
\begin{split}
&(\Sigma_1^{1/2}\Sigma_2\Sigma_1^{1/2})^{1/2} = \left(R_1S_1R_1^{\top}R_2S_2S_2R_2^{\top}R_1S_1R_1^{\top}\right)^{1/2} \\
&=\left(R_1\left(S_1R_1^{\top}R_2S_2\right)\left(S_1R_1^{\top}R_2S_2\right)^{\top}R_1^{\top}\right)^{1/2} \\
&=\left(R_1\left(\begin{array}{cc}
\frac{1}{4}w_1w_2(\cos \theta_1\cos \theta_2 + \sin \theta_1 \sin\theta_2) & 0 \\
0 & 0
\end{array}\right)\right.\\
&\left.\left(\begin{array}{cc}
\frac{1}{4}w_1w_2(\cos \theta_1\cos \theta_2 + \sin \theta_1 \sin\theta_2) & 0 \\
0 & 0
\end{array}\right)^{\top}R1^{\top}\right)^{1/2}\\
&=R_1\left(\begin{array}{cc}
\frac{1}{4}w_1w_2\cos \Delta\theta & 0 \\
0 & 0
\end{array}\right)R_1^{\top}
\end{split}
\end{equation}
Finally, the trace term is developed as
\begin{equation}
    \begin{split}
    \textbf{Tr}(...) &= \frac{w_1^2}{4} + \frac{w_2^2}{4} - \frac{1}{2}w_1w_2\cos \Delta\theta \\
    &=\frac{1}{4}||\Delta \mathbf{w}||^2
    \end{split}
\end{equation}
Hence, we get the simplified version of EGWD for a single edge pair as
\begin{equation}
    \begin{split}
        \mathbf{W_{12}} = ||\Delta \mathbf{\mu}||^2 + \frac{1}{4}||\Delta \mathbf{w}||^2
    \end{split}
\end{equation}

There are two constraints on the condition of OBox. First, the center points of each edge can be expressed by the 5-p vector $(x,y, w, h, \theta)$. Second, the parallel edges share identical covariance. 
We use $o$ to denote the box's center point, $\mu$ to denote the edge's center point, $\bf{w}$ to denote the vector corresponding to width and $\bf{h}$ to denote the vector corresponding to height.
By clockwise order, the center points of each edge can be written as
\begin{equation}
\label{eq:edge_center}
\begin{split}
    \mathbf\mu_1 &=o - \frac{\mathbf{h}}{2},
    \mathbf\mu_2 = o + \frac{\mathbf{w}}{2} \\
    \mathbf\mu_3 &=o + \frac{\mathbf{h}}{2},
    \mathbf\mu_4 = o - \frac{\mathbf{w}}{2} \\
\end{split}
\end{equation}
Then, the EGWD can be developed as
\begin{equation}
    \begin{split}
        \mathbf{W}(p,q) &= ||\Delta (o - \frac{\mathbf{h}}{2})||^2 + ||\Delta (o + \frac{\mathbf{w}}{2})||^2 \\
        &+ ||\Delta (o + \frac{\mathbf{h}}{2})||^2 + ||\Delta (o - \frac{\mathbf{w}}{2})||^2\\
        &+ \frac{1}{2}||\Delta \mathbf{w}||^2 + \frac{1}{2}||\Delta \mathbf{h}||^2\\
        &= 4 ||\Delta o||^2 + ||\Delta \mathbf{w}||^2 + ||\Delta \mathbf{h}||^2
    \end{split}
\end{equation}

\subsection{Full Derivation of Edge Dense Wasserstein Distance}
We denote the edge center point as $\mathbf{c}$, the edge vector as $\mathbf{w}$. For a pair of edge, suppose two sampled points $\mathbf{X}_1$, $\mathbf{X}_2$ from two edges can be denoted as 
\begin{equation}
    \mathbf{X}_i = \mathbf{c} + x\frac{\mathbf{w_i}}{2}
\end{equation}
where $x \in [-1, 1]$
Then the two edge's distance can be written as
\begin{equation}
\label{eq:derive1}
    \begin{split}
        \mathbf{W_{12}} &= \int_x p(x)||\Delta \mathbf{c} + x \frac{\Delta \mathbf{w}}{2}||^2 dx \\
        &=||\Delta\mathbf{c}||^2 + \int_x p(x)x^2\frac{||\Delta \mathbf{w}||^2}{4} dx + \int_x p(x)x\Delta\mathbf{c}\Delta\mathbf{w} dx
    \end{split}
\end{equation}
In this paper, we choose the probability density function $p(x)$ to be axial symmetric around the edge center. Thus, the last term of Eq-\ref{eq:derive1} is zero. Note that $\int_x p(x)x^2dx$ represents the variance of the distribution, which we can use $\sigma^2$ to denote it. Finally, the expression can be written as
\begin{equation}
    \label{eq:der2}
    \begin{split}
        \mathbf{W_{12}} = ||\Delta\mathbf{c}||^2 + \frac{\sigma^2}{4}||\Delta \mathbf{w}||^2
    \end{split}
\end{equation}
Combining Eq-\ref{eq:edge_center} and Eq-\ref{eq:der2}, we get the formulation for OBox
\begin{equation}
    \label{eq:OBox_edwd}
    \begin{split}
    \bf{W(p,q)} =  \bf 4||\Delta o||^2 &+ \left(\frac{1}{2} + \frac{\sigma_w^2}{2}\right)||\Delta \mathbf w||^2 \\
    &+ \left(\frac{1}{2} + \frac{\sigma_h^2}{2}\right)||\Delta \mathbf h||^2
    \end{split}
\end{equation}

{\small
\bibliographystyle{ieee_fullname}
}